\documentclass[sigconf]{acmart}
\usepackage{booktabs}
\usepackage[utf8]{inputenc}
\usepackage[T1]{fontenc}
\usepackage{hyperref,url}  
\usepackage{booktabs}
\usepackage{nicefrac}
\usepackage{microtype}
\usepackage{graphicx}
\usepackage{subcaption}
\usepackage{amsfonts,float, amsmath, mathtools, nicefrac}
\usepackage{color}
\usepackage{array,multirow,balance}
\usepackage{wrapfig}
\usepackage{tabularx}

\usepackage{array}
\newcolumntype{P}[1]{>{\centering\arraybackslash}p{#1}}

\newcommand{\ignore}[1]{}
\newcolumntype{Y}{>{\centering\arraybackslash}X}
\AtBeginDocument{%
  \providecommand\BibTeX{{%
    \normalfont B\kern-0.5em{\scshape i\kern-0.25em b}\kern-0.8em\TeX}}}

\usepackage{enumitem}
\begin{document}
\title{DeepBLE: Generalizing RSSI-based Localization Across Different Devices}
\author{Harsh Agarwal}
\email{harshaga@andrew.cmu.edu}
\affiliation{
  \institution{Carnegie Mellon University}
}

\author{Navyata Sanghvi}
\email{nsanghvi@andrew.cmu.edu}
\affiliation{
  \institution{Carnegie Mellon University}
}

\author{Vivek Roy}
\email{vroy@andrew.cmu.edu}
\affiliation{
  \institution{Carnegie Mellon University}
}

\author{Kris Kitani}
\email{kmkitani@andrew.cmu.edu}
\affiliation{
  \institution{Carnegie Mellon University}
}


\begin{abstract}
Accurate smartphone localization (< 1-meter error) for indoor navigation using only RSSI received from a set of BLE beacons remains a challenging problem, due to the inherent noise of RSSI measurements. To overcome the large variance in RSSI measurements, we propose a data-driven approach that uses a deep recurrent network, DeepBLE, to localize the smartphone using RSSI measured from multiple beacons in an environment. In particular, we focus on the ability of our approach to generalize across many smartphone brands (e.g., Apple, Samsung) and models (e.g., iPhone 8, S10). Towards this end, we collect a large-scale dataset of 15 hours of smartphone data, which consists of over 50,000 BLE beacon RSSI measurements collected from 47 beacons in a single building using 15 different popular smartphone models, along with precise 2D location annotations. Our experiments show that there is a very high variability of RSSI measurements across smartphone models (especially across brand), making it very difficult to apply supervised learning using only a subset smartphone models. To address this challenge, we propose a novel statistic similarity loss (SSL) which enables our model to generalize to unseen phones using a semi-supervised learning approach. For known phones, the iPhone XR achieves the best mean distance error of 0.84 meters. For unknown phones, the Huawei Mate20 Pro shows the greatest improvement, cutting error by over 38\% from 2.62 meters to 1.63 meters error using our semi-supervised adaptation method.
\end{abstract}


\maketitle

\section{Introduction}

Smartphone-based localization using Bluetooth Low Energy (BLE) beacons is utilized in various urban spaces, but its accuracy is severely limited by the large variance in Received Signal Strength Index (RSSI) measured across various smartphone models. Urban spaces such as hospitals, museums, shopping malls and airports utilize BLE beacons to offer location-specific services but their use has been largely limited to rough proximity sensing due to the instability of the RSSI measurement. Implementing more accurate forms of smartphone localization needed for applications such as way-finding (e.g., assisting blind people to navigate indoor spaces) is challenging due to this lack of consistency between smartphone receiver characteristics. In this work, we work towards developing a BLE beacon-based localization technique that can generalize to any smartphone model. 

\begin{figure}
    \centering
    \includegraphics[width=\linewidth]{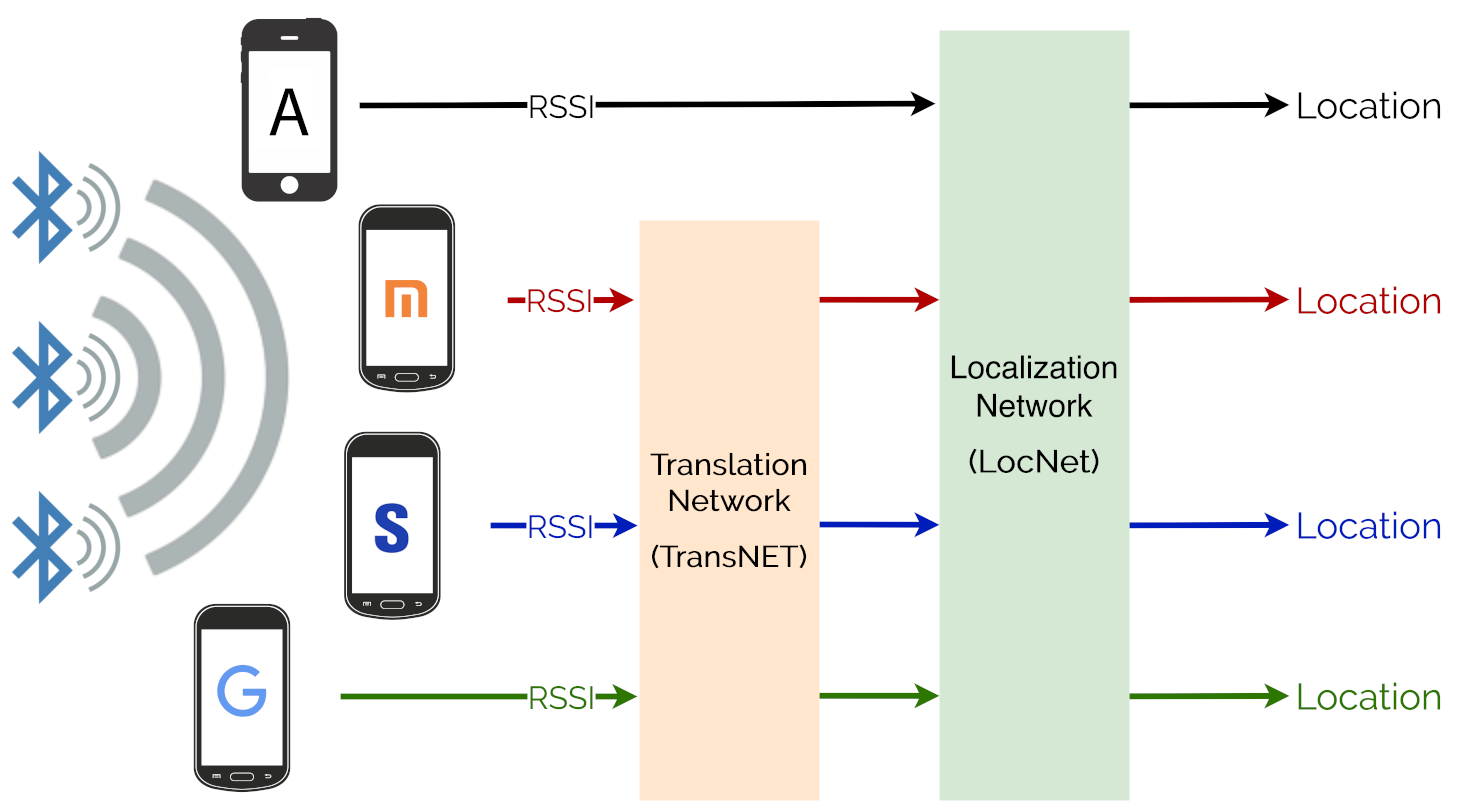}
    \caption{Proposed Approach: TransNet learns to transfer RSSI from any phone to RSSI from a reference phone we had during training. LocNet is the Localization engine to get positions given RSSI by BLE}
    \label{fig:my_label}
\end{figure}{}

The RSSI of BLE beacons measured by a smartphone can vary greatly for different brands and models due to a number of reasons. Hardware differences such as the form factor of the phone, receiver module, layout of the internal circuitry and antenna design can all affect how the RSSI is measured. Software factors such as built-in signal processing and data loss can also affect the RSSI measurements. Due to these differences in measurements, it is challenging to develop a single method that will work across multiple smartphones. Therefore, it is critical to develop localization methods that can adapt to changes in RSSI measurements across smartphones. 

Due to the differences in RSSI measurements across smartphones, fingerprinting based localization methods are a favorable approach. Whereas trilateration methods often require calibration of measurements to fit a known sensor model to compute location (e.g.,  log power drop off), fingerprinting methods bypass the need for calibration by storing a large database of measurements to index location. Fingerprinting also do not require a known parametric model and can capture complex non-linear changes in RSSI measurements that may be challenging to define. The benefit of robustness does come at the cost of a requiring a labor intensive data collection process, where RSSI measurements must be recorded at many locations in the environment prior to use. However, using indoor mapping technology such as robotic mapping systems, the cost of data collection can be reduced. 

In this work, we utilize a fingerprinting approach using a recurrent neural network which has the ability to generalize to any smartphone model. We first characterize the performance of a neural network-based localization system using a standard supervised learning approach and show how it fails to generalize to different smartphone models. We then propose a semi-supervise method for adapting the localization system by introducing a translation network which is able to take the RSSI measurements of an unseen smartphone and convert it to the RSSI measurements of a known smartphone.

To evaluate our method we collect a large-scale dataset of BLE beacon data collected using 15 popular smartphone models from Apple, Samsung, Google, Huawei and Xiaomi. The data is taken in a large building instrumented with 47 BLE beacons. 15 different users collected over 15 hours of RSSI measurement using various smartphones. The dataset consists of over 50,000 beacon RSSI measurements and is 10 times larger than existing datasets.

The contributions of this work are as follows: (1) we propose the concept of a translation function which maps the RSSI signals of unknown smartphones to the space of known smartphones; (2) we provide an empirical characterization of RSSI localization on 15 popular smartphones and quantify the large variance in RSSI measurements; (3) we present the largest BLE beacon RSSI dataset to enable thorough evaluation of localization techniques across multiple smartphone models.
\section{Related Work}

Various localization techniques based on RFID \cite{wang2013dude}, UWB radios \cite{uwb}, and ultrasound \cite{alps} have been developed. Recently, localization using Channel State Information (CSI) has been investigated to achieve a high accuracy (about 1m) \cite{rssi_to_csi, decimeterLevelLocalization,BLoc}. However, extracting the CSI in commodity smartphones is non-trivial. In contrast, RSSI information is available ubiquitously and is easy to retrieve. For this reason and relatively low infrastructure costs, using only RSSI from BLE or WiFi remain a popular choice for indoor localization \cite{gp_regression, pedometer_wifi, Subhan2011IndoorPI, faragher2015location}. \cite{BLE_vs_Wifi_oxford,he2015wi}, the authors conclude that BLE is much more suitable for the application of indoor positioning and navigation specifically.

RSSI-based localization can be categorized into two categories: fingerprinting and trilateration. Fingerprinting uses a database to store the RSSI strengths at various known reference points. At run time, this database is used to calculate the location. Both non-parametric method such as nearest neighbors, and parametric approaches such as neural networks can be used for this computation. However, as RSSI measurements must be collected for many locations in the environment, the process is often time-consuming and expensive. Moreover, as these methods are based on data of the RSSI  distribution obtained from a set of receivers, naively using these methods is ought to result in failure when the received RSSI distribution changes. Trilateration on the other hand uses parametric models of RSSI signal attenuation to form geometric constraints of the receiver with respect to a set of transmitters (or vice versa). Known methods are able to estimate position within 2 meters \cite{fingerprintingVStrialteration, liu2012push}. A variety of models have been proposed such as Gaussian models \cite{gp_regression}, Monte Carlo \cite{biswas2010wifi}, Bayesian \cite{bayesian_methods}, Hidden Markov Models \cite{markov_models}, and radio propagation models \cite{RADAR_2000}. Ring overlapping approaches have been proposed in the past \cite{vivekanandan2006concentric,range_free_sensing}, however, these methods tend to require additional calibration, both at the antenna and the phone level.This calibration process based on transmitters and receivers significantly inhibits their usage and generalization ability across different phones. Also modelling signal propagation methods for complex indoor environments is extremely difficult \cite{IP_problems} and lead to wrong estimates.

Hence, often for large spaces fingerprinting based RSSI localization remains a popular choice. \cite{classic_fingerprinting} Transmitters (some BLE, some WiFi) were deployed across a floor, and the ground truth locations were collected using an Active Bats system. Recently, \cite{murata2018smartphone} came up with adaptations to a classic particle filter (with an observation model as proposed in \cite{gp_regression}) in order to deploy beacon-based localization in a large scale building and achieving a localization error in the range of 2-3 m. However, the method involves a lot of parameter calibration for different scenarios, drawing questions on the generalization of the method across phones. 

With the advent of deep learning and machine learning, some methods have tried leveraging these parametric approximators for BLE localization. \cite{onefc_position, zhang2016deep, hall2019sdr, dabil} combine use of robust feature estimates from Deep networks, with classical machine learning,
algorithms like probabilistic KNNs and HMM's to come up with indoor positioning systems. . \cite{DLRNN-based_localization_wifi} proposed an end to end deep network performing RSSI based localization. Although the results have been published on WiFi RSSI signals, the end goal is very similar to what we are trying to achieve. Hence we use it as one of our baselines. Very recently, \cite{lstm_inverse_problem} tried solving the inverse problem of localizing beacon scanners, using RSSI strengths received. In the experiments section we compare a very similar network as proposed by \cite{lstm_inverse_problem} architecture to DeepBLE.  

Although there is some work on understanding variations of BLE signals across different smart phone, it is not very well understood. \cite{eval_ble_generalize} studies BLE technology on a variety smartphones models finding out key differences on per smart phone basis. However they do not test the BLE for Localization on different phones. Other works, like \cite{murata2018smartphone, Chen2017LocatingAT} evaluate their methods on additional phone models/devices, a rigorous test of generalization is missing. As we see BLE Localization systems  being deployed in real world scenarios, with almost 200+ smart phone models present this is inevitably a challenge that would inhibit the use case of RSSI based indoor localization methods in future.

In this work we take up this challenge, analysing BLE RSSI for multiple smartphones, creating a large scale dataset for 15 different smart phones and propose a localization engine that can achieve competitive results on all the phones, without explicitly training the neural network in a supervised fashion on each phone. 

\section{Problem Setup}
\label{sec:statement}

Consider an indoor environment with $B$ BLE beacons and $P$ different smartphone models (\emph{e.g.}, iPhone X, S10). Let the set of beacon indices be $\mathcal{B} = \{1...,B\}$, and the set of smartphone model indices be $\mathcal{P} = \{1,...,P\}$. 

At any time step $t$, smartphone $p \in \mathcal{P}$ is at location $\mathbf{x}^p_t \in \mathbb{R}^2$. 
The RSSI received by smartphone $p$ from beacon $b \in \mathcal{B}$  at time $t$ is $s^{p,b}_t \in \mathbb{R}_{\geq 0}$. The vector of RSSI received by smartphone $p$ from all $B$ beacons at time $t$ is $\mathbf{s}^p_t = [s_t^{p,1} \;\; s_t^{p,2} \;\; ... \;\; s_t^{p,B}]^T \in \mathbb{R}_{\geq 0}^{B}$. For a history of $H$ time steps, the matrix of RSSI received by smartphone $p$ from all beacons at times $\{t-(H-1), t-(H-2),...,t\}$ is $\mathbf{S}^p_t = [\mathbf{s}_{t-(H-1)}^p \;\; \mathbf{s}_{t-(H-2)}^p \;\; ... \;\; \mathbf{s}_{t}^p] \in \mathbb{R}_{\geq 0}^{B\times H}$

Consider the labeled dataset $\mathcal{D}_l = \{\langle\mathbf{S}^p_t, \mathbf{x}^p_t\rangle \;| \; {p \in \mathcal{P}, \; t \in \mathcal{T}_l}\}$. Each element of $\mathcal{D}_l$ is a tuple, consisting of the matrix of RSSI received by a smartphone $p$, and its corresponding location at some time $t \in \mathcal{T}_l$. 

Consider the unlabeled dataset  $\mathcal{D}_u = \{\mathbf{S}^p_t \;|\; {p \in \mathcal{P}, \; t \in \mathcal{T}_u}\}$. Each element of $\mathcal{D}_u$ is the matrix of RSSI received by a smartphone $p$ at some time $t \in \mathcal{T}_u$.  

Let there be an unknown function $f: \mathbb{R}_{\geq 0}^{B\times H} \rightarrow \mathbb{R}^2$ mapping the matrix of RSSI received by any smartphone to the location of the smartphone at that time, i.e. $f(\mathbf{S}) = \mathbf{x}$. In the problem of indoor localization, our goal is to approximate this function. In this paper, we consider the following scenarios:
\begin{enumerate}[wide, labelindent=0pt, topsep=0pt]
    \item We consider the ideal scenario where we have access to {\bf labeled data} from all smartphones $\mathcal{D}_l$. 
    \item We consider the more realistic scenario where we only have access to {\bf limited labeled data} from a subset of smartphone models. Let $\mathcal{P}'$ be a subset of smartphone models, i.e., $\mathcal{P}' \subset \mathcal{P}$. Then, the corresponding subset of labeled data is $\mathcal{D}'_l = \{\langle\mathbf{S}^p_t, \mathbf{x}^p_t\rangle \;| \; {p \in \mathcal{P}', \; t \in \mathcal{T}_l}\} \subset \mathcal{D}_l$. We consider the situation where we only have access to $\mathcal{D}'_l$.
    \item Lastly, we consider the scenario where we have access to {\bf limited labeled data} from a subset of smartphone models, as well as {\bf limited unlabeled data} from a subset of unseen smartphone models. Let $\mathcal{Q}'$ be a subset of smartphones such that $\mathcal{Q}' \in \{\mathcal{P}\setminus\mathcal{P}'\}$. Then, the corresponding subset of unlabeled data is $\mathcal{D}'_u = \{\mathbf{S}^p_t \;| \; {p \in \mathcal{Q}', \; t \in \mathcal{T}_u}\} \subset \mathcal{D}_u$. We consider the situation where we only have access to $\mathcal{D}'_l$ and $\mathcal{D}'_u$.
\end{enumerate}

In each scenario, our goal is to approximate the function $f$ from the available data $\mathcal{D}_{\text{train}}$, such that our approximation $\hat f$ provides accurate localization of any smartphone in a known environment. Specifically, \; \; scenario 1:\; $\mathcal{D}_{\text{train}} = \mathcal{D}_l$, \;\; scenario 2:\; $\mathcal{D}_{\text{train}} = \mathcal{D}'_l$,  \;\;scenario 3:\; $\mathcal{D}_{\text{train}} = \mathcal{D}'_l \cup \mathcal{D}'_u$.

\section{Proposed Approach}
\label{sec:approach}

As discussed in Section \ref{sec:statement}, our goal is to construct an approximation $\hat f$ to the localization function $f: \mathbb{R}_{\geq 0}^{B\times H} \rightarrow \mathbb{R}^2$, which maps the matrix of RSSI received by any smartphone to the location of the smartphone at that time. In each of the three scenarios we consider, the localization function $\hat f$ is approximated with a deep neural network. For each scenario, we propose different  paramterizations of $\hat{f}$, objectives, and training procedures based on the data available for training. We refer to our proposed methods collectively as {\bf DeepBLE}.

\subsection{Scenario 1: Labeled Data for All Smartphone Models}

We develop our first DeepBLE method for learning the localization function $\hat f$ where we are given labeled data from all phone models for training, \emph{i.e.}, $\mathcal{D}_{\text{train}} = \mathcal{D}_l$. While this scenario is ideal, it is often unrealistic as it requires that we obtain the ground truth location for every possible smartphone model for training.

\subsubsection{Localization Network (\texttt{LocNet}):} The fully supervised localization network $\hat f$ is modeled as a 2 layer LSTM followed by two fully connected layers. We call the localization network, \texttt{LocNet}. At time $t$, it takes as input $\mathbf{S}^p_t$, i.e., the beacon readings received by smartphone $p$ in the last $H = 5$ seconds, and outputs the location $\mathbf{x}^p_t$. During training, \texttt{LocNet} learns to correctly interpret temporal beacon information, e.g.: how fast signals are changing, patterns of oscillation in signals, etc., and maps it to a location $\mathbf{x}$.

\subsubsection{Optimization:}  We define the following objective function:
\begin{align}    
    \label{eqn:Dl}
    \min\limits_{\hat f}\;\;\;  &\mathbb{E}_{\langle\mathbf{S},\mathbf{x}\rangle \sim \mathcal{D}_{l}}\left[ \Big|\Big|\hat f(\mathbf{S}) - \mathbf{x}\Big|\Big|^2_2 \right]
\end{align}
This is the standard empirical risk minimization objective for supervised learning, where we learn the minimizer $\hat{f}$ over the training data for all phone model $\mathcal{P}$.

\subsection{Scenario 2: Limited Labeled Data}

We now develop our second DeepBLE method for a more realistic scenario, where we have access to limited labeled data from a subset of smartphone models $\mathcal{P}' \subset \mathcal{P}$, i.e., $\mathcal{D}_{\text{train}} = \mathcal{D}'_l$. Each model included in the subset $\mathcal{P}'$ belongs to a brand. For example, $\mathcal{P}'$ may include iPhone 7 and iPhone 8 from the Apple brand as well as Pixel 3 and Pixel 4 from the Google brand. In this example, $\mathcal{P}'$ does not include any models from Samsung, Xiaomi or Huawei brands.

If we learn a localization function according to optimization \ref{eqn:Dl} using only the limited labeled data $\mathcal{D}'_l$, we will overfit to the smartphone brands included in $\mathcal{P}'$ since the training data is not representative of all smartphone brands. In order to avoid overfitting, we propose the use of a BLE signal translation function which maps RSSI measured by any smartphone to that of a smartphone brand which is known, \emph{i.e.,} one which was part of the labeled training dataset.

Consider a brand-specific localization function $h: \mathbb{R}_{\geq 0}^{B\times H} \rightarrow \mathbb{R}^2$, which maps the matrix of RSSI measured by smartphone models of a known brand to the smartphone's location. Let the set of smartphone models of that known brand be indicated by $\mathcal{J} \subset \mathcal{P'}$.

Now consider a BLE signal translation function $g: \mathbb{R}_{\geq 0}^{B\times H} \rightarrow \mathbb{R}_{\geq 0}^{B\times H}$, which takes as input the matrix of RSSI received by any smartphone model $p \in \mathcal{P}$ and transforms it to a corresponding matrix of RSSI measured by a model $p' \in \mathcal{J}$ from the known smartphone brand, at the same location. Then, $f(\mathbf{S}) = h(g(\mathbf{S}))$ is the true localization function, where $\mathbf{S}$ is the matrix of RSSI received by a smartphone model $p \in \mathcal{P}$. Our goal is to learn an estimate of the localization function $\hat f = \hat h(\hat g)$.

For this scenario, we define a new loss function with the following components:
\begin{enumerate}[wide, labelindent=0pt,topsep=0pt]
    \item \textbf{Localization loss:} $\mathcal{L}_{\text{loc}} = \big|\big|\hat h(\hat g(\mathbf{S}_t)) - \mathbf{x}_t\big|\big|^2_2$. This quantifies deviation of estimates of our localization function from ground truth label $\mathbf{x}$ and is identical to the loss function in scenario 1.
    
    \item \textbf{Position smoothness loss:} This quantifies the distance between consecutive location predictions, $\mathcal{L}_{\text{ps}} =  \big|\big|\hat h(\hat g(\mathbf{S}_t)) - \hat h(\hat g(\mathbf{S}_{t-1}))\big|\big|^2_2$. This term acts as a regularization term to ensure that estimated motion is smooth.
    
    \item \textbf{Statistic similarity loss (SSL):} Intuitively, we know that beacons that are near each other will often be `seen' together by the smartphone Bluetooth receiver. This means that beacon RSSI measurements are correlated and that the outputs of the translation function $g$ should also be correlated. Formally, consider the subset of models $\mathcal{J} \subset \mathcal{P}'$ of the known brand . Labeled data from models of this brand is given by $\mathcal{D}'_{J} = \{\langle\mathbf{s}^p_t, \mathbf{x}^p_t\rangle \;| \; {p \in \mathcal{J}, \; t \in \mathcal{T}_l}\}$. We use the brand-specific data $\mathcal{D}'_J$ to infer brand-specific correlation statistics.
    
    Now, consider a matrix $\mathbf{M} \in \mathbb{R}_{\geq0}^{B\times B}$ of statistics derived from data $\mathcal{D}'_J$. Each entry of $\mathbf{M}$ is described as:
    \begin{align*}
        \mathbf{M}_{ij} = \mathbb{E}_{\langle\mathbf{s},\mathbf{x}\rangle \sim \mathcal{D}'_J}[\mathbf{s}_{j} \;|\; \mathbf{s}_{i} > 0],
    \end{align*}
    where $\mathbf{M}_{ij}$ is the expected RSSI from the j$^{\text{th}}$ beacon when the RSSI from the i$^{\text{th}}$ beacon is detected.
    
    We use the inferred statistics to quantify the deviation of the output of $\hat g$ from the expected statistics of smartphone models $\in \mathcal{J}$. Our proposed statistic similarity loss (SSL) is defined by:
    \begin{align*}
        \mathcal{L}_{\text{ssl}} = \sum\limits^H_{t=1} \sum\limits^B_{i=1}\sum\limits^B_{j=1} w_{it} d_{ijt}
    \end{align*}
    where:
    \begin{align*}
        w_{it} &= \exp\left(\frac{\left(\mathbf{S}_{it} - \mathbf{M}_{ii}\right)}{\tau}\right), \text{\;\;\;and}\\
        d_{ijt} &= \begin{cases}
        ||\;\mathbf{M}_{ij} - \hat g(\mathbf{S})_{jt}\; ||_1 & \text{if \;\; }\mathbf{M}_{ij} - \hat g(\mathbf{S})_{jt} < 0 \\
        \left(\mathbf{M}_{ij} - \hat g(\mathbf{S})_{jt}\right)^2 & \text{if \;\; } \mathbf{M}_{ij} - \hat g(\mathbf{S})_{jt} \geq 0
        \end{cases}
    \end{align*}
    and $\tau$ is a hyperparameter. In our experiments, $\tau = 0.1$. Recall that the output of $\hat g$ is a matrix of translated RSSI and $\hat g(\mathbf{S})_{jt}$ denotes the entry corresponding to the j$^\text{th}$ beacon at time index $t$.
    
  Our function $\hat g$ must map the matrix of RSSI received by any smartphone model to the corresponding RSSI received by a known smartphone model $\in \mathcal{J}$. Our proposed loss SSL quantifies the deviation of the output of $\hat g$ from the expected statistics of smartphone models $\in \mathcal{J}$. Specifically, $\mathcal{L}_{ssl}$ provides an exponential weighting ($w_{it}$) to any deviation ($d_{ijt}$) from the expected RSSI of the j$^{\text{th}}$ beacon when the i$^{\text{th}}$ beacon is detected, according to pairwise statistics $\mathbf{M}_{ij}$ of known smartphone models $\in \mathcal{J}$. When $\hat g$ is learned correctly the statistical deviation of the translated output signals should be low.

    \item \textbf{Temporal smoothness loss:} One of the major problems we observed in lower-end smartphone models is lost RSSI information where beacons randomly measure zero RSSI. In order to address this problem, we introduce a temporal smoothness loss that ensures that the output of the translation function $\hat{g}$ does not randomly drop RSSI,
    \begin{align*}
        \mathcal{L}_{\text{ts}} = \sum\limits^{H-1}_{t=1} \sum\limits^B_{i=1} || \;\hat g(\mathbf{S})_{i,t} - \hat g(\mathbf{S})_{i,t+1}\;||_1.
    \end{align*}
    $\hat g(\mathbf{S})_{i,t}$ is the entry corresponding to the i$^\text{th}$ beacon at time index $t$.
\end{enumerate}

\subsubsection{Optimization:} We propose a modified optimization with weights $w_\text{loc}$, $w_\text{ps}$, $w_\text{ssl}$, $w_\text{ts}$ on each term respectively:
\begin{align}
    \label{eqn:dlprime}
    \min\limits_{\hat h,\;\hat g} \mathbb{E}_{\langle\mathbf{S},\mathbf{x}\rangle \sim \mathcal{D}'_{l}} \big[w_{\text{loc}}\mathcal{L}_{\text{loc}} + w_{\text{ps}}\mathcal{L}_{\text{ps}} + w_{\text{ssl}}\mathcal{L}_{\text{ssl}} + w_{\text{ts}}\mathcal{L}_{\text{ts}}\;\big]
\end{align}

By training $\hat h$ and $\hat g$ to minimize this objective, we expect generalization of localization performance to unseen smartphones.
The localization network $\hat h$ has the same architecture as \texttt{LocNet} detailed in Scenario 1. We now describe the signal translation network $\hat g$.

\subsubsection{Signal Translation Network (\texttt{TransNet}):}

Our signal translation function $\hat g$ transforms the matrix of RSSI measured by a phone $p \in \mathcal{P}$, to the matrix of RSSI measured by a model of a known brand in the same location. Intuitively, we expect the difference in RSSI measured by different phones at the same location to be bounded. Therefore, we constrain $\hat g$ to output a correction to its input. Formally, our signal translation function is a neural network of the form $\hat g(\mathbf{S}) = \textrm{ReLU}(\hat r(\mathbf{S}) + \mathbf{S})$. We call this network \texttt{TransNet}. The correction $\hat r(\mathbf{S})$ is a 6-layered 1-D convolutional autoencoder, composed of 3 encoder layers and 3 decoder layers with ReLU activations. 

\subsection{Scenario 3: Limited Labeled and Unlabeled Data}

We now describe our third DeepBLE approach for the scenario in which RSSI measurements for new smartphones (not contained in the labeled data) can be collected as people walk around the indoor environment. However, this data is not annotated with the smartphone's true location, \emph{i.e.,} the data is unlabeled. In this scenario, we have access to limited labeled data $\mathcal{D}'_l$ from smartphone models $\mathcal{P}' \subset \mathcal{P}$, as well as limited unlabeled data $\mathcal{D}'_u$ from other smartphone models $\mathcal{Q}' \in \{\mathcal{P}\setminus\mathcal{P}'\}$.

\subsubsection{Optimization:} Like the previous scenario, we consider the estimation of $\hat h$ and $\hat g$, so that the localization function $\hat f = \hat h(\hat g)$. Intuitively, using the additional unlabeled data, we can further tune the translation function $\hat g$ to deal with changes in the input distribution. For example, some smartphones may exhibit more RSSI signal dropping. Training $\hat g$ and $\hat h$ to deal with such changes further equips our localization function $\hat f$ to generalize to unseen smartphones. While the previous two scenarios involved a supervised approach to learning $\hat f$, in this scenario, we describe a {\bf semi-supervised} approach:
\begin{enumerate}[wide, labelwidth=!, labelindent=0pt]
    \item Using $\mathcal{D}'_l$, we first perform optimization (\ref{eqn:dlprime}) in a supervised manner, to obtain initial estimates $\hat h$ and $\hat g$.
    
    \item Once we have initial estimates of $\hat h$ and $\hat g$, we use both labeled data $\mathcal{D}'_l$ and unlabeled data $\mathcal{D}'_u$ from new smartphones $\mathcal{Q}'$ to improve our estimates. Notice that, in optimization (\ref{eqn:dlprime}), only the localization loss $\mathcal{L}_{loc}$ depends on the label $\mathbf{x}$. Thus, we define a modified objective function for the unlabeled data by adding loss functions that do not depend on the label. We use weights $w_\text{loc}$, $w_\text{ps}$, $w_\text{ssl}$, $w_\text{ts}$ as before, as well as a weight $w_u$ on the expected loss over the unlabeled data:
    \begin{align}
        \label{eqn:dldu}
        &\;\; \min\limits_{\hat h,\;\hat g}\;\; & \mathbb{E}_{\langle\mathbf{S},\mathbf{x}\rangle \sim \mathcal{D}'_{l}} \big[& w_{\text{loc}}\mathcal{L}_{\text{loc}} + w_{\text{ps}}\mathcal{L}_{\text{ps}} + w_{\text{ssl}}\mathcal{L}_{\text{ssl}} + w_{\text{ts}}\mathcal{L}_{\text{ts}}\;\big] + \nonumber\\
        && w_{u}\; \mathbb{E}_{\mathbf{x} \sim \mathcal{D}'_{u}} \big[& w_{\text{ps}}\mathcal{L}_{\text{ps}} + w_{\text{ssl}}\mathcal{L}_{\text{ssl}} + w_{\text{ts}}\mathcal{L}_{\text{ts}}\;\big]
    \end{align}
\end{enumerate}

By training $\hat h$ and $\hat g$ to minimize this objective, we expect better generalization of localization performance to unseen smartphones. The localization network $\hat h$ and signal translation network $\hat g$ have the same architecture as \texttt{LocNet} and \texttt{TransNet} detailed in the previous scenarios.

\section{Dataset}

\begin{figure*}[ht!]
\begin{subfigure}{0.28\linewidth}
    \centering
    \includegraphics[width=\linewidth]{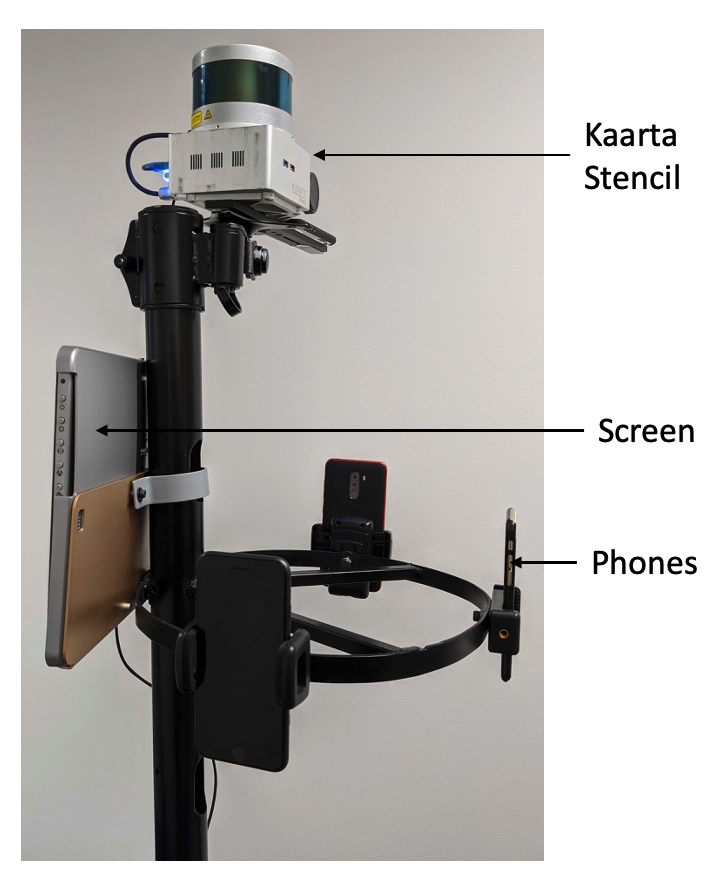}
    \caption{Data Collection Setup}
\end{subfigure}
\begin{subfigure}{0.6\linewidth}
    \centering
    \includegraphics[width=\linewidth]{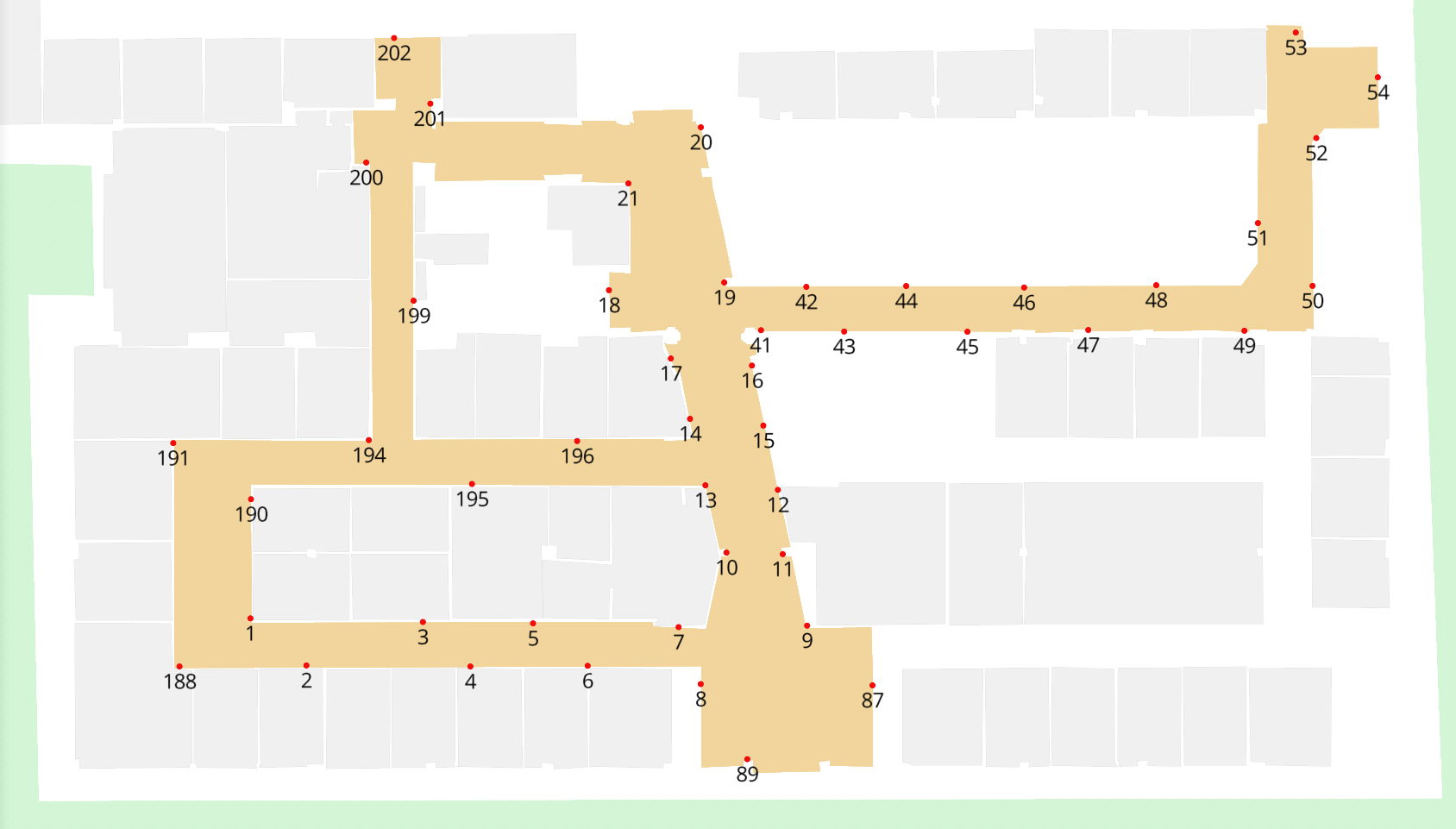}
    \caption{Beacon layout map showing minor ID of every beacon}
\end{subfigure}
\caption{ The colored region is the area we collect data in. The red dots show the beacons, along with the minor ids}
\label{fig:rig}
\end{figure*}

\begin{table}[ht]
\setlength\tabcolsep{1pt}
\SMALL
\def\arraystretch{1.1}
\begin{center}
\begin{tabularx}{\linewidth}{X c Y X}
\toprule
 \textbf{Model} & \textbf{Price} & \textbf{OS} & \textbf{Processor} \\
  \midrule
  Apple iPhone XR & \$600 & iOS 13 &  A12 Bionic  \\
  Apple iPhone 8 & \$450 & iOS 13 &  A11 Bionic  \\
  Apple iPhone 7 & \$190 & iOS 13 &  A10 Fusion  \\
  Samsung Galaxy S10 & \$700 & Android 10 & SM8150 SD-855*  \\
  Samsung Galaxy Note9 & \$600 & Android 9 & SDM845 SD-845*  \\
  Samsung Galaxy A50 & \$300 & Android 9 & Exynos9610  \\
  Google Pixel 4XL & \$900 & Android 10 & SM8150 SD-855*  \\
  Google Pixel 3XL & \$550 & Android 10 &  SDM845 SD-845*  \\
  Google Pixel 3aXL & \$400 & Android 10 &  SDM670 SD-670*  \\
  Xiaomi Mi 9 & \$435 & Android 9 & SM8150 SD-855*  \\
  Xiaomi Mi 9T Pro & \$360 & Android 9 & SM8150 SD-855*  \\
  Xiaomi Redmi Note 8 & \$180 & Android 9 & Mediatek Helio G90T  \\
  Huawei Mate 20 Pro & \$550 & Android 9 & HiSilicon Kirin 980  \\
  Huawei Honor 20 Pro & \$340 & Android 9 & HiSilicon Kirin 980  \\
  Huawei Honor View 20 & \$290 & Android 9 & HiSilicon Kirin 980  \\
  \bottomrule
\end{tabularx}
\end{center}
\caption{Specs of Smartphone models used as BLE Receivers. *Qualcomm Snapdragon }
\label{table:phone_price}
\vspace{-5mm}
\end{table}

Indoor localization using RSSI has been gaining momentum in the last few years, but the scarcity of large scale public datasets on BLE Localization has been a problem for researchers working toward this problem. To the best of our knowledge, \cite{mohammadi2017semi} is the only publicly available BLE RSSI dataset. The dataset was collected using iPhone6S, receiving BLE RSSI from 13 different iBeacons deployed in the Waldo Library of Western Michigan University. The dataset contains two sub-datasets: a labeled dataset (1420 instances) and an unlabeled dataset (5191 instances). As we try to analyze how varying phone types, change RSSI signals, the data cannot be used for our analysis. 

So towards this end, we collected a large scale BLE dataset with about 54K RSSI fingerprint samples along with fine-grained location information, for 15 smartphone models. The setup was deployed across a university floor spanning 2000 sq. m. in area. The dataset comprises of RSSI data ranging from Kontakt.io beacons for fifteen different phones from 5 major brands which include Apple, Samsung, Pixel, Huawei, and Xiaomi. In this section, we describe the data collection procedure in detail.

\subsection{Overview}
We designed a rig, as shown in \ref{fig:rig}, which can hold three phones and a Kaarta Stencil on top to get precise ground truth locations. Study in \cite{BLE_vs_Wifi_oxford}, suggests BLE data changes significantly with orientations, so having three facing directions helps us in augmenting the BLE data substantially at every single location. About 50 Kontakt.io beacons were installed all across the first floor of a university, as shown in \ref{fig:rig}. One of the factors that influence noise in RSSI systems is multipath fading, which often makes localization using BLE a lot more difficult. Therefore, the complex structure of the building makes it a perfect testing ground for our purpose.

\subsection{Obtaining Ground truth Locations}
Typically fingerprinting is performed standing at specific reference locations, collecting the RSSI and then using the BLE data at these reference locations to localize. However, as the data collection procedure is static, we cannot expect the methods to work very well when the robot/person is moving. However, collecting ground-truth requires high-end infrastructure in the building, making it extremely infeasible for scaling such methods. Stencil by Kaarta is a stand-alone, lightweight and low-cost system delivering the integrated power of mapping and real-time position estimation with a precision of $\pm 30$mm.  The device is based on the scientific work \cite{zhang2014loam} and depends on LIDAR, vision and IMU data for localization.  The system uses Velodyne VLP-16 connected to a low-cost MEMS IMU and a processing computer running Robotic Operating System (ROS) for real-time six DoF mapping and localization. A 10-Hz scan frequency is used for the data capture \ignore{VLP-16 has a $360^\circ$ field of view with a $30^{\circ}$ azimuthal opening using 16 scan lines. The stencil tilt angle is recommended to be within the $\pm 15^{\circ}$ envelope. The progress of the mapping can be monitored on-line via an external monitor attached via an HDMI cable.}\cite{lehtola2017comparison}. For robust estimates, we use the multi-modal localization engine, which gives us position estimates using data captured via both camera and LIDAR.

\subsection{Transmitters - BLE Beacons}

For our analysis and experiments, we used beacons manufactured by Kontakt.io. Some of the key factors and parameters of BLE beacons that can significantly influence localization performance are \textit{Broadcasting Power}, \textit{Advertising Interval}, and \textit{Measured Power}. In our setup these values are -12dBm, 100ms and -77 dBm respectively. We refer readers to \cite{BLE_101} for detailed analysis and impact of each of these terms. 

\subsection{Receivers - Commodity Smartphones}

We used fifteen different smartphone models from 5 notable brands, with their price ranging between \$250 to \$ 900 (Table \ref{table:phone_price}) as receivers. Broadly, choice in every brand comprises of one lower-end phone, one middle-end phone and one high-end smartphone. The detail and specs for every phone are consolidated in Table \ref{table:phone_price}. Note that every phone used has BLE version 5, except iPhone 7, which is using a version 4.2. 

\subsection{API for accessing BLE Beacon RSSI data}

iOS (on iPhones) has an inbuilt API to log the beacon data being sensed. However, we note that these readings are processed, and the maximum frequency with which one can access the data is 1 Hz. For Android, we built an Android Beacon logging library that logs beacon scans at a consistent rate of 1Hz with zero packet loss in processing.

\subsection{Time Synchronization}
 
We match the data recorded from the different phones with the ground truth from the stencil using timestamps recorded at both ends. In order to do this accurately, the time on the devices need to be synced. We perform NTP time sync on both the stencil and the phones at the beginning of every run using the same NTP server\footnote{NTP server used: time.google.com}. We performed empirical tests using IMU based calibration and observed the NTP sync to be precise to a few microseconds.

\subsection{Data Collection Procedure}

We place Kontakt.io beacons across the floor of a university building, as shown in Figure \ref{fig:rig}(b). The rig is designed to hold three phones (Figure \ref{fig:rig}(a)). We group the phones by brand and collect around 60 mins of data for each brand. The colored region Figure \ref{fig:rig}(b) is the regions where we perform localization. To make sure the comparison between the phones is fair, and the beacon signals have not changed drastically between different brands,on any given day we collected data for all brands by the same person. This process is repeated on multiple days to collect the complete data set. The data has been collected, during the after-hours in the evening, so that noise due to external factors is subsidized, and we can focus mainly on change in RSSI distributions introduced by varying smartphone models. For different runs, we ensure that the gait style and walking pattern does not vary a lot between brands. As we collect different runs of data, we observe that as the frequency of BLE data reception is low (typically 1 Hz), the BLE data received does not change drastically from person to person for normal walking speeds (1-1.2m/s). Expecting absolutely no change between different runs is not possible. However, the best efforts were made to collect the data to be as similar as possible to a person walking with a phone in hand, with almost all factors we can control remaining the same except the smartphone models.

\subsection{Train Test Validation Split}

We refer to the collected Dataset as $\mathcal{D}_c$ which we split into the train set ($\mathcal{D}_{c,\textrm{train}}$), validation set ($\mathcal{D}_{c,\textrm{val}}$), and test set ($\mathcal{D}_{\textrm{test}}$). Each of these sets are independent runs, with no overlapping samples. We note that through out our analysis, $\mathcal{D}_{\textrm{test}}$ remains untouched and is used only for final evaluation.

\section{BLE RSSI Analysis}
\label{sec:ble_analysis}

\begin{table}[tb]
\setlength\tabcolsep{3pt}
\def\arraystretch{1}
\begin{center}
\begin{tabular}{ c c c c c c c } 
\toprule
\textbf{Env.} $\rightarrow$& \multicolumn{3}{c}{\textbf{No Interference}} & \multicolumn{3}{c}{\textbf{W/ Interference}} \\
\cmidrule(lr){1-1} \cmidrule(lr){2-4} \cmidrule(lr){5-7}
{\textbf{Phone}$\downarrow$} & {\textbf{Mean}} & {\textbf{Var}} & {\textbf{SnR}} & {\textbf{Mean}} & {\textbf{Var}} & {\textbf{SnR}} \\
  \cmidrule(lr){1-1} \cmidrule(lr){2-4} \cmidrule(lr){5-7} 
    iPhone 7 & 27.61 & 7.18 & 3.83 &26.80 & 4.07 & 6.59 \\
    iPhone 8 & 30.20 & 1.73 & 17.49& 27.52 & 8.25 & 3.34  \\
    iPhone XR & 29.17 & 2.26 & 12.87& 27.93 & 0.37 & 76.47  \\
    Galaxy S10 & 33.04 & 8.09 & 4.09& 20.87 & 14.77 & 1.41\\
    Galaxy Note9 & 30.30 & 2.86 & 10.58 &22.35 & 40.92 & 0.55 \\
    Galaxy A50 & 30.83 & 17.93 & 1.72 & 26.80 & 19.03 & 1.41 \\
    Pixel 4XL & 24.46 & 11.47 & 2.13 & 11.83 & 41.93 & 0.28 \\
    Pixel 3XL & 30.35 & 0.49 & 62.22& 16.41 & 42.72 & 0.38 \\
    Pixel 3aXL & 28.98 & 3.20 & 9.07 & 20.61 & 50.37 & 0.41\\
    Mi 9 & 27.39 & 45.90 & 0.60 & 19.11 & 41.77 & 0.46 \\
    Mi 9T Pro & 30.39 & 55.79 & 0.54& 19.12 & 62.34 & 0.31 \\
    Redmi Note 8 & 32.33 & 4.35 & 7.43 & 22.50 & 43.95 & 0.51\\
    Mate 20 Pro & 34.67 & 1.39 & 24.88& 26.33 & 21.87 & 1.20 \\
    Honor 20 Pro & 28.04 & 11.91 & 2.35& 21.07 & 35.63 & 0.59 \\
    Honor View 20 & 26.09 & 3.99 & 6.53 & 25.17 & 7.06 & 3.57 \\
  \bottomrule
\end{tabular}
\end{center}
\caption{Per phone signal statistics of one beacon in absence (No interference) and in presence (W/ interference) of other beacons. RSSI distribution changes significantly across phones.}
\label{table:SNR_table}
\vspace{-5mm}
\end{table}

\begin{figure}
\begin{subfigure}{.49\linewidth}
    \centering
    \includegraphics[width=\linewidth]{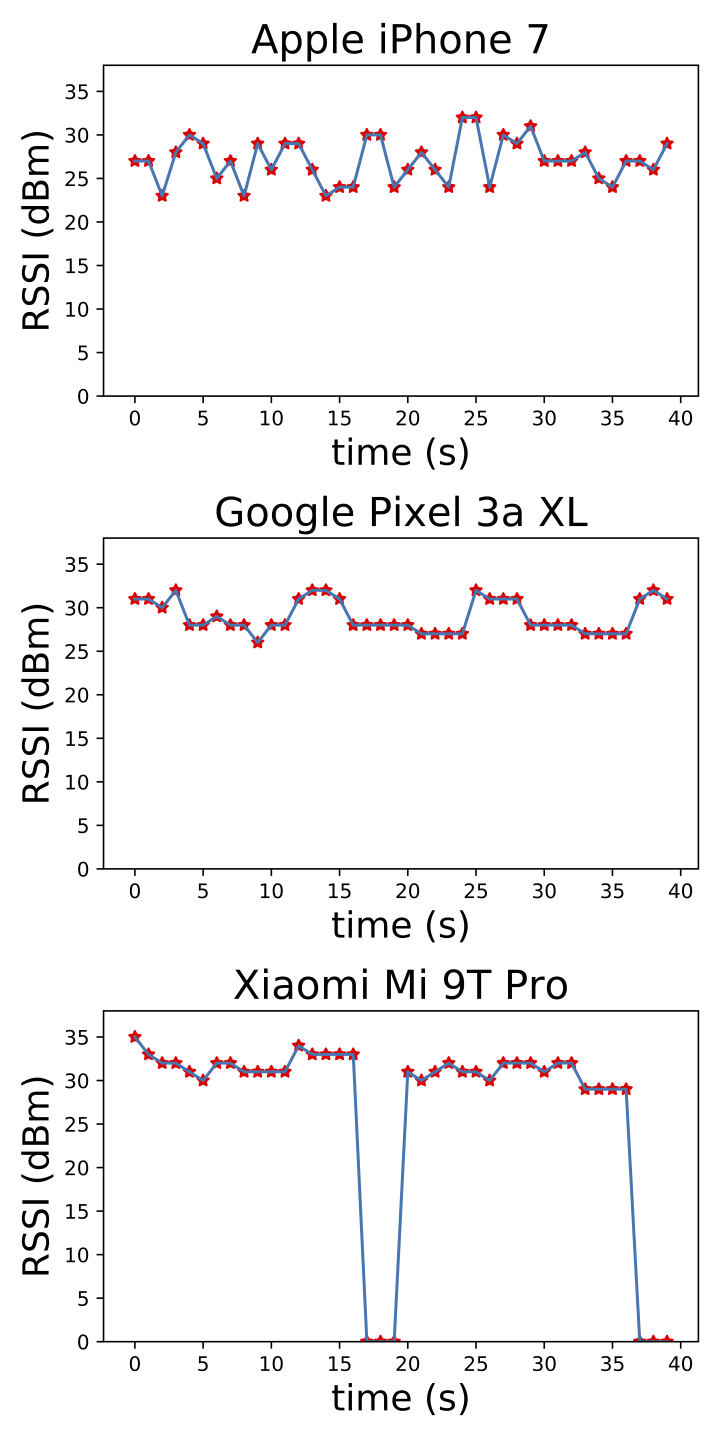}
    \caption{Isolation (Exp. 1)}
\end{subfigure}
\begin{subfigure}{.49\linewidth}
    \centering
    \includegraphics[width=\linewidth]{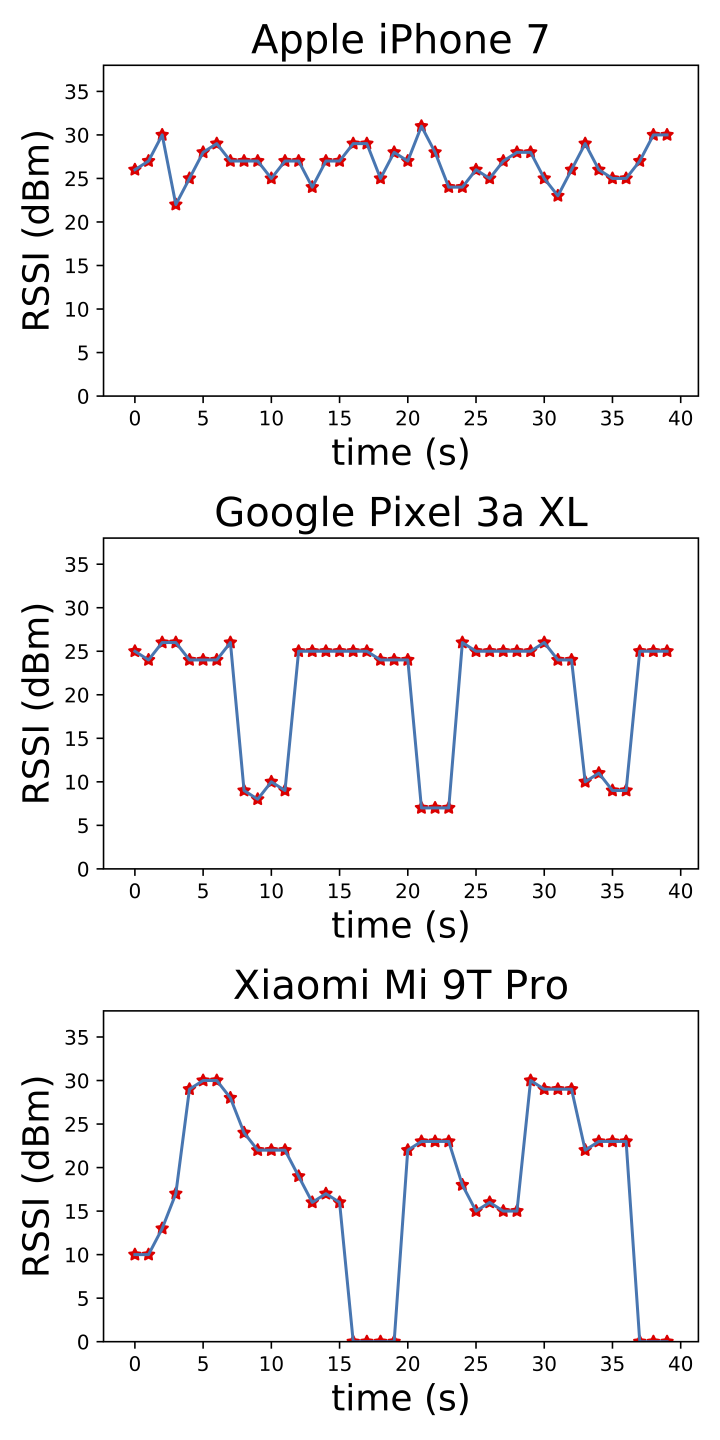}
    \caption{Building (Exp. 2)}
\end{subfigure}
\caption{RSSI vs. Time plots for the two experimental setups show that having multiple beacons in the environment has little impact on the nature of the signal from iPhone, some impact on the Pixel phone, but major impact in case of the Xiaomi phone. We see significantly higher instability in Experiment 2}
\label{fig:exp12}
\end{figure}

\subsection{RSSI without Interference}

In an isolated environment, we consider one beacon and one smartphone at a time. The BLE beacon and smartphone are kept at a distance of 1m, and RSSI data is collected for 60 seconds for each phone sequentially. We calculate the mean and the variance of the RSSI signal for each phone using the signal strengths registered during the 1-minute interval (Table \ref{table:SNR_table} and Figure \ref{fig:exp12}) Although the mean RSSI being registered across different phones is approximately the same, the variance changes drastically across phones.  For particular phones, signal drops are observed. For instance, for Mi 9T Pro, we observe the beacon was not registered for 3 seconds continuously. 

 \subsection{RSSI with Interference}
 
We repeat a similar experimental setup, with the same beacon, but in the presence of multiple other beacons to act as sources of interference. According to \cite{IP_problems}, we know multi-path fading, and interference due to multiple beacons can lead to a significant change in RSSI behavior. The numbers are reported in \ref{table:SNR_table}, under the "W/Interference" heading. The experiments were performed in the building, where the dataset has been collected. We observe in addition to the variance, in this case, the mean RSSI changes significantly across phones, contrary to what we observed in the no interference setting. Also, we note that the variance (noise) is not directly related to what we observed in the absence of other sources of interference. For instance, Pixel3XL records a significant variance in this setup, and iPhone XR shows a remarkably stable reading, Figure (\ref{fig:exp12}). The differences we observe can be due to multiple factors, which include multi-path fading and interference due to multiple beacons. Empirically we conclude, that the behaviour of RSSI distribution can change significantly across phones in the presence of other sources of interference.

\subsection{RSSI Receiver Failure Statistics}

A troubling phenomenon that we encountered were frequent occurrence of receiver failure where all RSSI measurements become zero at random time instances. We compute the following metrics to quantify receiver failure: (1) \textit{Non Zero Mean RSSI} - The average RSSI strength that a phone detects for RSSI measurements greater than -100 dBm; (2) \textit{Receiver Failure} - Percentage of samples in the train data when no beacons were detected; (3) \textit{Dead time} - Mean duration of such Receiver failure incidents. These metrics obtained for each phone are reported in Table \ref{table:Stats_train}. Note that receiver failures happen about 15\% of the time for Xiaomi models and can last roughly from 1 to 3 seconds. These statistics suggest that the translation network that we implement needs to be able to handle significant receiver failures where all beacons RSSI measurements are completely missing.

\begin{table}[tb]
\setlength\tabcolsep{3pt}
\def\arraystretch{1}
\begin{center}
\begin{tabular}{ c c c c} 
\toprule
\textbf{Phone} & \textbf{Mean} & \textbf{Receiver } & \textbf{Dead } \\
\textbf{Name} & \textbf{RSSI(dBm)} & \textbf{Failure(\%) } & \textbf{time(s)} \\
  \cmidrule{1-1} \cmidrule{2-2} \cmidrule{3-3} \cmidrule{4-4}
    iPhone 7 & 12.09 &0.00 & 0.00 \\
    iPhone 8 & 12.84. & 0.00 & 0.00 \\
    iPhone XR & 11.64& 0.00 & 0.00 \\
    Mate 20 Pro & 9.28& 9.98 & 1.45 \\
    Honor 20 Pro &8.28 & 8.52 & 1.51 \\
    View 20 & 7.389 & 14.96 & 1.55 \\
    Mi9 & 8.27 & 15.18 & 3.20 \\
    Mi9TPro & 10.03. & 15.11 & 3.19 \\
    Redmi Note 8 &. 10.03  &7.88 & 1.00 \\
    Pixel 3aXL & 9.68 &0.00 & 0.00 \\
    Pixel 3XL & 10.00&0.00 & 0.00 \\
    Pixel 4XL &8.63 & 0.00 & 0.00 \\
    Samsung S10 &13.18 & 0.16 & 1.73 \\
    Samsung A50 & 10.25& 0.17 & 1.31 \\
    Samsung Note9 &9.25& 0.16 & 1.32 \\
  \bottomrule
\end{tabular}
\end{center}
\caption{RSSI Receiver Failure Statistics}
\label{table:Stats_train}
\vspace{-8mm}
\end{table}

\begin{figure}[tb]
    \centering
    \includegraphics[width=0.8\linewidth]{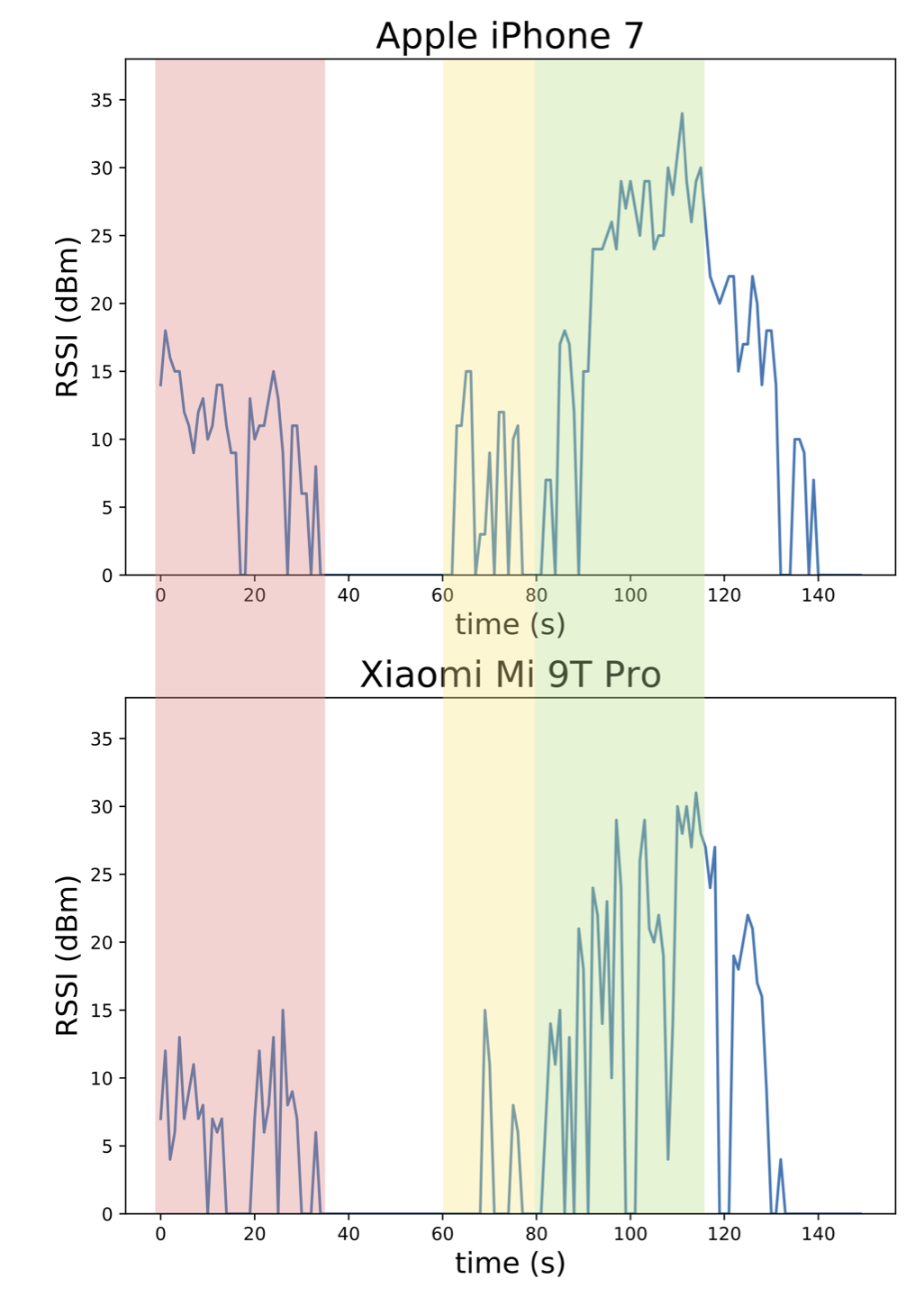}
    \caption{RSSI by iPhone 7 vs. RSSI by Xiaomi}
    \label{fig:exp3}
\end{figure}

To provide a more qualitative visualization of the dead time issue, we had a subject walk with two phones (one in each hand), across the building recording RSSI from 47 possible beacons on both phones. We plot the RSSI vs. Time plot for a particular beacon ($b=11$) for the Apple iPhone 7 and Xiaomi Mi9T Pro shown in Figure \ref{fig:exp3}. Notice that the Mi9T Pro has many instances of signal dropping (no RSSI measurements) over time. In contrast, the iPhone 7 has a strong RSSI signal throughout.

\section{Experiments: DeepBLE Methods}
\label{sec:experiments}

\begin{table*}[ht]
\setlength\tabcolsep{3pt}
\setlength{\arrayrulewidth}{0.25mm}
\def\arraystretch{1}
\begin{center}
\begin{tabular}{ l| r l |r l r l r l |r l r l } 
\toprule
\multicolumn{1}{l|}{{\textbf{Scenario} $\rightarrow$}} &\multicolumn{2}{c|}{\bf Scenario 1} &
\multicolumn{6}{c|}{\bf Scenario 2} &
\multicolumn{4}{c}{\bf Scenario 3}  \\
\cmidrule{2-3}\cmidrule{4-9} \cmidrule{10-13}
& \multicolumn{2}{c|}{\multirow{2}{*}{\bf Proposed}} & \multicolumn{2}{c}{\multirow{2}{*}{\bf Baseline 1}} & \multicolumn{2}{c}{\multirow{2}{*}{\bf Baseline 2}} & \multicolumn{2}{c|}{\multirow{2}{*}{\bf Proposed}} & \multicolumn{2}{c}{\bf Proposed} & \multicolumn{2}{c}{\bf Proposed}\\
& & & & & & & & & \multicolumn{2}{c}{(Huawei)} & \multicolumn{2}{c}{(Xiaomi)}\\
\midrule
{Test Phone} & {Mean} & {\;\;Std\;\;} & {Mean} & {\;\;Std\;\;} & {Mean} & {\;\;Std\;\;} & {Mean} & {\;\;Std\;\;} & {Mean} & {\;\;Std\;\;} & {Mean} & {\;\;Std\;\;} \\
\toprule 

iPhone 7 &0.95 &{0.65}  &0.99 &0.83 &0.92 &0.83 &\textbf{0.91} & \textbf{0.75} &0.95 &0.88 & \textbf{0.94} &  \textbf{0.82} \\
iPhone 8 &1.09 &  {0.73} &1.13 &0.95 &1.16 &0.99 & \textbf{1.06} & \textbf{0.76} &1.11 & \textbf{0.83} &  \textbf{1.07} & \textbf{0.83} \\
iPhone XR &0.92 &0.60 &1.05 &0.83 &0.95 &0.75  & \textbf{0.84} &\textbf{0.60} &0.89 &0.80 & \textbf{0.86} & \textbf{0.63} \\

Mate 20 Pro &1.96 &1.61 &4.19 &4.25 &2.62 &2.45  & \textbf{1.76} & \textbf{1.61} & \textbf{1.63} &1.32 & \textbf{1.63} &\textbf{1.31} \\
Honor 20 Pro &  {1.92} &1.46 &3.14 &2.19 &2.36 &1.77  & \textbf{1.95} & \textbf{1.42} & 1.99 &\textbf{1.33} & \textbf{1.92} &1.34 \\
View 20 &1.74 &1.40 &3.15 &3.31 &2.20 &2.04  &\textbf{1.64} & \textbf{1.59} &1.66 &1.32 &\textbf{1.64} &\textbf{1.29} \\

Xiaomi Mi 9 &  {1.53} &1.20 &3.82 &5.48 &2.39 &3.30  & \textbf{1.69} & \textbf{1.57} &1.65 &1.31 & \textbf{1.58} &\textbf{1.11} \\
Xiaomi Mi 9T Pro &1.77 &  {1.14} &3.78 &4.98 &2.71 &3.35  & \textbf{1.78} & \textbf{1.41} &1.84 &1.45 &\textbf{1.68} & \textbf{1.20} \\
Xiaomi Redmi Note 8\;\;\; &  {1.50} &1.17 &2.25 &2.01 &1.76 &1.57  & \textbf{1.62} & \textbf{1.16} & \textbf{1.61} &1.09 & \textbf{1.61} &\textbf{1.06} \\

Pixel 3a XL &  {1.05} &  {0.83} &1.51 &1.30 &1.25 &1.12  & \textbf{1.24} & \textbf{1.07} &1.22 &1.07 & \textbf{1.21} & \textbf{1.06} \\
Pixel 3XL &1.21 &0.96 &2.04 &1.56 &1.54 &1.29  & \textbf{1.22} & \textbf{0.95} &\textbf{1.17} &0.96 &1.18 &\textbf{0.94} \\
Pixel 4XL &  {1.21} &  {0.83} &2.13 &1.64 &1.53 &1.28  &  \textbf{1.36} & \textbf{1.01} & \textbf{1.39} & \textbf{1.07} &1.40 &1.08 \\

Samsung S10 &1.44 &1.15 &1.80 &1.41 &1.58 &1.35  & \textbf{1.31} & \textbf{1.05} & 1.30 &1.02 &\textbf{1.23} &\textbf{0.93} \\
Samsung A50 &1.49 &  {0.98} &2.11 &1.68 &1.62 &1.34  & \textbf{1.42} & \textbf{1.30} &\textbf{1.36} & \textbf{1.22} &1.42 &1.26 \\
Samsung Note9 &1.14 &0.87 &1.67 &1.23 &1.27 &0.98  & \textbf{1.13} & \textbf{0.82} & \textbf{1.15} &\textbf{0.81} & \textbf{1.15} &0.82 \\

\midrule[\heavyrulewidth]
Overall Average &1.37 &1.11 &2.23 &2.76 &1.67 &1.83  &\textbf{1.37} &\textbf{1.20} &1.37 &1.15 &\textbf{1.34} &\textbf{1.09} \\

\bottomrule
\end{tabular}
\vspace{1mm}
\caption{Mean and std. deviation of absolute localization error for all methods and scenarios. All numbers are in meters(m). Numbers in bold indicate best performance within each scenario. Numbers in color indicate best performance overall.}
\label{table:ablation}
\end{center}
\end{table*}

\begin{table}[ht]
\setlength\tabcolsep{3pt}
\setlength{\arrayrulewidth}{0.25mm}
\def\arraystretch{1}
\begin{center}
\begin{tabular}{l|c|ccc|cc } 
\toprule
{\multirow{3}{*}{\bf Stat. $\downarrow$}} & {\bf Scen. 1} &
\multicolumn{3}{c|}{\bf Scen. 2} &
\multicolumn{2}{c}{\bf Scen. 3}  \\
\cmidrule{2-7}
& {\multirow{2}{*}{\bf Prop.}} &  {\multirow{2}{*}{\bf Base. 1}} &  {\multirow{2}{*}{\bf Base. 2}} &  {\multirow{2}{*}{\bf Prop.}} &  {\bf Prop.} &  {\bf Prop.}\\
& & & & & {(Hua.)} &  {(Xia.)}\\
\toprule
Mean & 1.37 & 2.23  & 1.67 & {1.37}  & 1.37 & \textbf{1.34} \\
Std. Dev. & 1.11 & 2.76 & 1.83  & {1.20} & 1.15 & \textbf{1.09} \\
Median & 1.07 & 1.47 & 1.15 & \textbf{1.02} & 1.06 & 1.04 \\
90 \%ile & 2.68 & 4.55 & 3.49 & 2.76 & 2.76 & \textbf{2.66} \\
Max. & 12.28 & 37.58 & 29.67 & 19.72 & 13.19 & \textbf{11.82} \\

\bottomrule
\end{tabular}
\vspace{1mm}
\caption{Various statistics of absolute localization error over all test data for all methods and scenarios. All numbers are in meters (m). Numbers in bold indicate best performance over all methods.}
\label{table:overall}
\end{center}
\end{table}

\begin{table*}[ht]
\setlength\tabcolsep{3pt}
\def\arraystretch{1}
\begin{center}
\begin{tabular}{ c c c c c c c c c c c c c } 
\toprule
\multicolumn{1}{c}{\textbf{Method} $\rightarrow$} &  
\multicolumn{2}{c}{KNN Reg} &
\multicolumn{2}{c}{Bayesian} &
\multicolumn{2}{c}{KRR+KF} &
\multicolumn{2}{c}{DL-RNN} &
\multicolumn{2}{c}{Proposed(Sup)} &
\multicolumn{2}{c}{Proposed (Semi)} \\
\cmidrule(lr){1-1} \cmidrule(lr){2-3} \cmidrule(lr){4-5} \cmidrule(lr){6-7} \cmidrule(lr){8-9} \cmidrule(lr){10-11} \cmidrule(lr){12-13} 
\textbf{Test on $\downarrow$} & {Mean} & {Std} & {Mean} & {Std} & {Mean} & {Std} & {Mean} & {Std} & {Mean} & {Std} & {Mean} & {Std} \\
\cmidrule(lr){1-1} \cmidrule(lr){2-3} \cmidrule(lr){4-5} \cmidrule(lr){6-7} \cmidrule(lr){8-9} \cmidrule(lr){10-13} 
iPhone &1.1917 &0.9298 & 1.5424 & 1.2267 &1.7575 &1.0770 &1.0205 &0.7712 &\textbf{0.9409} &\textbf{0.7163} &0.9572 &0.7741 \\
Huawei &6.3579 &9.595 & 4.8951 & 6.3216 &3.0021 &2.3075 &2.2437 &2.2014 &1.7900 &1.5515 &\textbf{1.7370} &\textbf{1.3260} \\
Xiaomi &4.1781 &7.2262 &3.5959 &4.2674 &2.7569 &1.9125 &2.4360 &3.0378 &1.7014 &1.3918 &\textbf{1.6284} &\textbf{1.1280} \\
Pixel &1.9082 &1.5561 & 2.1457 & 1.7167 &1.8955 &1.1816 &1.4347 &1.1638 &1.2769 &\textbf{1.0183} &\textbf{1.2687} &1.0341 \\
Samsung &2.0139 &2.5525 & 2.0843 & 2.3115 &2.0767 &1.2390 &1.5282 &1.1301 &1.2898 &1.0808 &\textbf{1.2732} &\textbf{1.0310} \\
\midrule[\heavyrulewidth]
Average &3.0157 &5.7190 & 2.7724 & 3.8090 &2.2640 &1.6644 &1.6826 &1.8650 &1.3700 &1.2080 &\textbf{1.3462} &\textbf{1.0979} \\
Median &\multicolumn{2}{c}{1.5230} &\multicolumn{2}{c}{1.7505} &\multicolumn{2}{c}{1.8884} &\multicolumn{2}{c}{1.2067} &\multicolumn{2}{c}{\textbf{1.0258}} &\multicolumn{2}{c}{1.0436} \\
90\% ile &\multicolumn{2}{c}{5.2514} &\multicolumn{2}{c}{5.5866} &\multicolumn{2}{c}{4.2660} &\multicolumn{2}{c}{3.2968} &\multicolumn{2}{c}{2.7689} &\multicolumn{2}{c}{\textbf{2.6601}} \\
Max &\multicolumn{2}{c}{52.5572} &\multicolumn{2}{c}{48.3013} &\multicolumn{2}{c}{17.2977} &\multicolumn{2}{c}{31.9299} &\multicolumn{2}{c}{19.7231} &\multicolumn{2}{c}{\textbf{11.8294}} \\
\bottomrule
\end{tabular}
\caption{Evaluation results for different methods when trained using data from iPhone and Samsung phones}
\label{table:evaluation}
\end{center}
\end{table*}

In this section, we present experiments performed using the DeepBLE methods developed in Section \ref{sec:approach}. We implement all methods using PyTorch \cite{pytorch}. Before detailing experiments for different scenarios, we describe our augmentations to the collected data $\mathcal{D}_{\text{c,train}}$ and the resulting labeled data set $\mathcal{D}_l$. We also provide details of the evaluation metrics used.

\subsubsection{Data augmentation:} Based on the analysis of RSSI across different smartphone models (Section \ref{sec:ble_analysis}), we augment the collected train data $\mathcal{D}_{c,\textrm{train}}$ as follows:
 
 \begin{itemize}[wide, labelwidth=!, labelindent=0pt]
     \item \textbf{Scaling the RSSI strengths}: Dependence of beacons on batteries often leads to a decrease in RSSI transmitted by a beacon with time. To replicate that effect, we scale down the BLE input $\mathbf{S}$ at multiple scales. 
    \item \textbf{Imitating packet loss}: Given only three channels for transmission in BLE, Packet Loss is often observed when there are too many beacons in an environment. We imitate such behaviour by adding random signal drops in the input signals in a constrained fashion.
     \item \textbf{Random Noise}: As the beacon data being received changes with scene dynamics, we add Gaussian white noise sampled from $\mathcal{N}(0,5)$ to our BLE input.
 \end{itemize}
 
We refer to the augmented data as $\mathcal{D}_A$. Thus, the labeled data from all smartphone brands is $\mathcal{D}_l = \mathcal{D}_{c,\textrm{train}} \cup \mathcal{D}_A$.

\subsubsection{Evaluation:} In all scenarios, our methods are evaluated using $\mathcal{D}_{\text{test}}$. All reported statistics are with respect to absolute localization error (AE) of the learned localization function $\hat f$ on test sample $\langle\mathbf{S}, \mathbf{x}\rangle \sim \mathcal{D}_{\text{test}}$\;:
\begin{align*}
    \text{AE}(\mathbf{S},\mathbf{x}) \;=\; \Big|\Big| \hat f(\mathbf{S}) - \mathbf{x} \Big|\Big|_2
\end{align*}
We now discuss our methods and results for each scenario in detail. For all methods, Table \ref{table:ablation} represents mean and standard deviation of absolute localization error over test data from each smartphone model, as well as over all test data $\mathcal{D}_{\text{test}}$. For all methods, Table \ref{table:overall} represents mean, standard deviation, median, 90 percentile and maximum absolute localization error over all test data $\mathcal{D}_{\text{test}}$. 
\subsection{Scenario 1: Labeled Data for All Smartphone Models}

As discussed in Section \ref{sec:approach}, in our first scenario, we have labeled data from all smartphone models, \emph{i.e.,} $D_{\text{train}} = \mathcal{D}_l$. We train the localization network \texttt{LocNet} for objective (\ref{eqn:Dl}) in Section \ref{sec:approach}, using the Adam optimizer for 50 epochs, with a learning rate $= 10^{-4}$ and batch size $=$ 256 samples.

\subsubsection{Results:} We report mean and standard deviation of absolute localization error over test data in Table \ref{table:ablation}. Note, we do not have any numbers in bold here, since we do not compare our proposed method with any other method in this scenario. In this ideal scenario, since we have labeled data for all phones, performance on all phones is very good, with an average AE of {\bf 1.37 m} over all test data. Despite the high amounts of receiver failure for smartphone models from Huawei and Xiaomi brands (Section \ref{sec:ble_analysis}), AE for each of these models is {\bf less than 2 m}. However, as noted before, it is unrealistic to assume that we can performing fingerprinting for all smartphone models.

\subsection{Scenario 2: Limited Labelled data}
In our second scenario, we have only limited labeled data $\mathcal{D}_{\text{train}} = \mathcal{D}'_l$ from a subset of smartphone models $\mathcal{P}' \subset \mathcal{P}$. In our experiments with this scenario, $\mathcal{P}'$ includes models of the two most popular smartphone brands \footnote{as reported by the IDC - Worldwide Quarterly Mobile Phone Tracker} - Apple and Samsung, \emph{i.e.,} $\{$iPhone 7, iPhone 8, iPhone XR, Samsung S10, Samsung A50, Samsung Note 9$\}$. Our signal translation function $\hat g$ learns to transform RSSI measured by any smartphone model to that measured by the Apple brand, \emph{i.e.,} $\mathcal{J} \subset \mathcal{P}'$ includes smartphone models from the Apple brand $\{$iPhone 7, iPhone8, iPhone XR$\}$. 
As detailed in Section \ref{sec:approach}, our localization function in this scenario is $\hat f = \hat h(\hat g)$, where $\hat h$ is \texttt{LocNet} and $\hat g$ is \texttt{TransNet}. We train $\hat f$ for objective (\ref{eqn:dlprime}) in Section \ref{sec:approach}, using the Adam optimizer for 50 epochs, with a learning rate $= 10^{-4}$ and batch size $=$ 256 samples. Appropriate weights for each loss term are obtained using validation data, which was a randomly chosen independent run from the training data for each phone. we use the validation data from only $\mathcal{D}'_l$

\subsubsection{Baselines:} We compare our method against the following baselines:

\begin{enumerate}[wide, labelwidth=!, labelindent=0pt, topsep=0pt]
    \item \textit{Baseline 1}: Similar to the first scenario, we consider the situation where our localization function $\hat f$ is \texttt{LocNet} only. We limit the available labeled data to that from smartphone models of the Apple brand only, \emph{i.e.,} $\mathcal{P}'$ includes \{iPhone7, iPhone8, iPhone XR\}. We train \texttt{LocNet} according to optimization (\ref{eqn:Dl}) in Section \ref{sec:approach}, using labeled data from the Apple brand. We include this baseline to verify our hypothesis that data augmentation alone is not enough to generalize well to unseen phones. We expect our proposed method to outperform this method due to the inclusion of labeled data from an additional smartphone brand.
    \item \textit{Baseline 2}: Again, we consider the situation where our localization function $\hat f$ is \texttt{LocNet} only. However, for a stronger baseline, we consider that we have access to the same data as our proposed method, i.e., $\mathcal{P}'$ includes models from the Apple and Samsung brands. We train \texttt{LocNet} according to optimization (\ref{eqn:Dl}) in Section \ref{sec:approach} using labeled data from Apple and Samsung brands. We include this baseline to verify our hypothesis that simply training \texttt{LocNet} using limited labeled data will lead $\hat f$ to overfit to the known phone brands. We expect our proposed method to outperform this method due to the inclusion of a signal translation function $\hat g$. This is because $\hat g$ explicitly learns to translate RSSI measured by a (possibly unknown) smartphone to that measured by a known brand, using our novel statistic similarity loss (SSL). This enables $\hat f = \hat h(\hat g)$ to generalize better to unseen phones. 
\end{enumerate}

\subsubsection{Results:} From Table \ref{table:ablation}, we see that our proposed method significantly outperforms baselines in terms of both mean and standard deviation of AE over all test data, as well as over test data from individual smartphone models. Moreover, our proposed approach performs as well as the fully supervised approach from the first scenario, achieving an average AE of {\bf 1.37 m}, despite having access to only limited labeled data. Importantly, this demonstrates its ability to generalize well to unseen smartphone models. 

While Baseline 2 performs significantly better than Baseline 1, it performs poorly on unseen smartphone brands Huawei and Xiaomi. By comparison, our proposed approach consistently gives mean AE of {\bf less than 2 m} for models of these brands. Moreover, from Table \ref{table:overall}, Baseline 2 shows a high maximum error of $\approx 30 m$. This supports our hypothesis that training with limited data leads \texttt{LocNet} to overfit to known smartphones. Our proposed approach avoids this problem by training an explicit signal translation function $\hat g$ using our novel SSL loss.

Baseline 1 performs very well on iPhone models, with mean and standard deviation of AE within 1 m. However, its performance is significantly worse on other smartphone brands, especially Huawei and Xiaomi. Moreover, from Table \ref{table:overall}, the maximum and 90 percentile errors are the highest in the table. Baseline 1 overfits to a larger degree than Baseline 2, supporting our hypothesis that data augmentation alone is not enough to generalize well to unseen phones.

\subsection{Scenario 3: Limited Labeled and Unlabeled Data}

In our third scenario, we have limited labeled data $ \mathcal{D}'_l$ from a subset of smartphone models $\mathcal{P}' \subset \mathcal{P}$. Additionally, we have unlabeled data $ \mathcal{D}'_u$ from other smartphone models $\mathcal{Q}' \subset \{\mathcal{P}\setminus\mathcal{P}'\}$. Our training data is thus $\mathcal{D}_{\text{train}} = \mathcal{D}'_l \cup \mathcal{D}'_u$. Similar to the previous scenario, in our experiments with this scenario, $\mathcal{P}'$ includes models of Apple and Samsung brands. Our signal translation function $\hat g$ learns to transform RSSI measured by any smartphone model to that measured by the Apple brand. We consider $\mathcal{Q}'$ to include smartphone models of the Huawei brand or the Xiaomi brand. As detailed in Section \ref{sec:approach}, our localization function in this scenario is $\hat f = \hat h(\hat g)$, where $\hat h$ is \texttt{LocNet} and $\hat g$ is \texttt{TransNet}. In this scenario,
\begin{enumerate}[wide, labelwidth=!, labelindent=0pt, topsep=0pt]
    \item We train $\hat f$ for objective (\ref{eqn:dlprime}) in Section \ref{sec:approach}, using the Adam optimizer for 50 epochs, with a learning rate $= 10^{-4}$ and batch size $=$ 256 samples. Appropriate weights for each loss term are obtained using validation data.
    \item We then train $\hat f$ for the {\bf semi-supervised} objective (\ref{eqn:dldu}) in Section \ref{sec:approach}, using the Adam optimizer for 20 epochs, with a learning rate $= 10^{-5}$ and batch size $=$ 256 samples.
\end{enumerate}

\subsubsection{Results:} From Table \ref{table:ablation}, we see that our proposed method trained in a semi-supervised fashion using unlabeled data from the Xiaomi brand leads to better test performance for Xiaomi and Huawei smartphones, when compared to our proposed method in scenario 2. Moreover, our proposed method {\bf outperforms the fully supervised method} in scenario 1, with a mean AE of {\bf 1.34 m} and standard deviation AE of {\bf 1.09 m} over all test data. Importantly, this result is achieved despite only having access to limited labeled data, demonstrating our proposed method's ability to generalize well to unseen phones. From Table \ref{table:overall}, our proposed method leads to the best (or nearly the best) performance on all metrics over all the test data. 


\begin{figure*}
    \centering
    \includegraphics[width=0.75\linewidth]{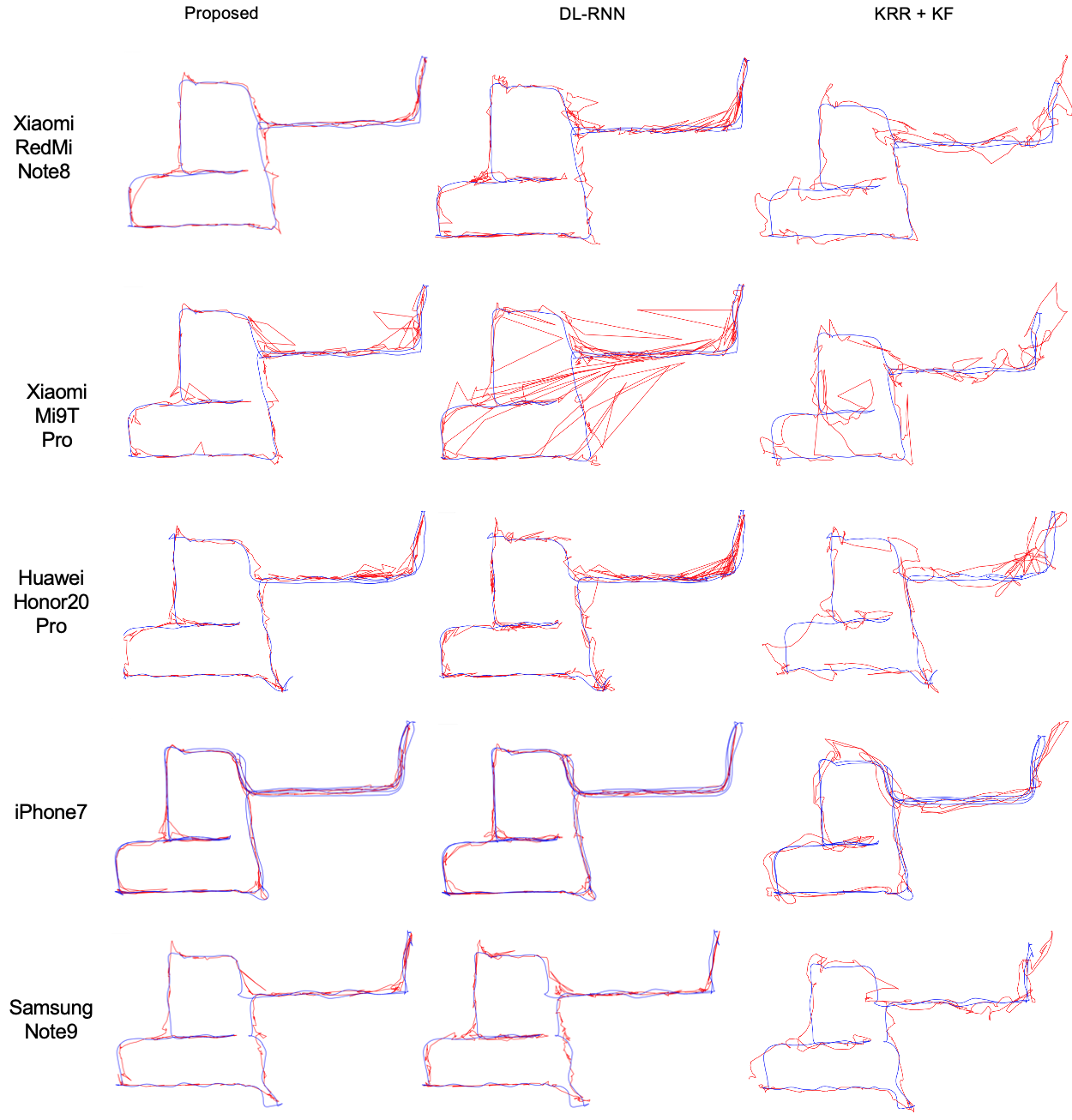}
    \caption{Visualization of Localization Performance comparing proposed approach to other approaches.Blue denotes groundtruth, Red denotes predictions}
    \label{fig:my_label}
\end{figure*}

\begin{figure*}
\begin{subfigure}{.4\linewidth}
    \centering
    \includegraphics[width=\linewidth]{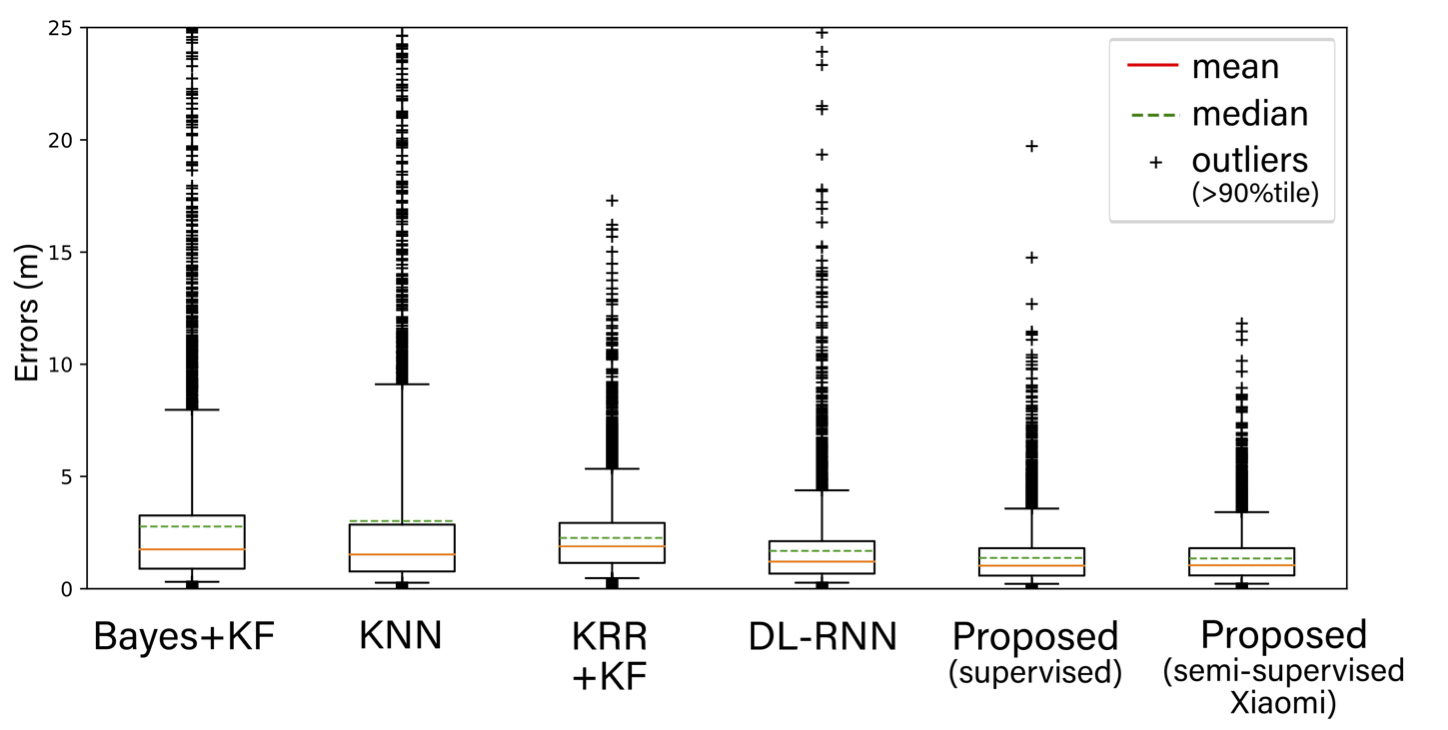}
    \caption{Meam, Median and 90\%ile Errors }
\end{subfigure}
\begin{subfigure}{.29\linewidth}
    \centering
    \includegraphics[width=\linewidth]{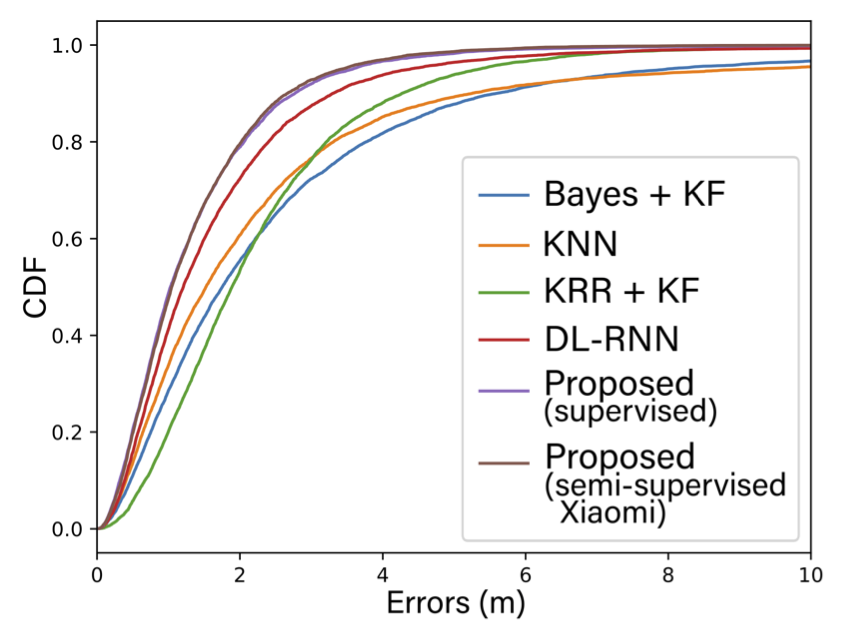}
    \caption{Distance Error CDF comparing different methods}
\end{subfigure}
\caption{Evaluation}
\label{fig:box_cdf}
\end{figure*}

\section{Experiments: Common Localization Methods}

We verify our proposed approach and compare it with some of the most commonly used and state of the art methods in RSSI based Localization. Particularly, the performance of DeepBLE has been compared to the 1. K-Nearest Neighbours (KNN)\cite{RADAR_2000} 2. Bayesian Estimation \cite{Chen2013BayesianFF} 3. NavCog (clustering based Ridge Regression with a Kalman Filter)\cite{murata2018smartphone} and 4. DL-RNN \cite{DLRNN-based_localization_wifi}. For KNN, we perform a weighted KNN regression using k=10, and a weighing function inversely proportional to the distance from each considered neighbour. As \cite{murata2018smartphone} uses information from multiple sensors, we make a comparison to the observation model in the particle filter proposed to make a fair comparison to our proposed method as it's observation model depends on only RSSI based Localization. \cite{DLRNN-based_localization_wifi} although based on WiFi, solves the same problem of regressing locations directly using deep learning given only RSSI signals, and can be easily deployed for a BLE RSSI setting. Hence we use that as a baseline as well. For all the methods we use a time history of 5 seconds of BLE beacon RSSI to make a fair comparison between them. 

To evaluate each of the methods mentioned above, we calculate the mean and standard deviation of the absolute distance error, as used in \ref{sec:experiments}. We calculate this metric on per brand basis as well as across all the phones. We report the median, 90\%ile errors and the max for each of the above methods to evaluate the robustness of each method.  The comparisons for all the methods have been made under Scenario 2: Access to limited labelled data, as none of these methods can use the unlabelled data in some way.  

The numbers obtained are shown in Table \ref{table:evaluation} and the visual comparison on the test data for competitive approaches are shown in Figure \ref{fig:my_label}. The proposed approach outperforms all the other methods in terms of the mean and standard deviation of distance errors. More importantly, the disparity between the localization performance shown by phones of different brands is significantly reduced. KNN, which is one of the most commonly used methods, the errors shoot up drastically as we switch phone brands indicating its susceptibility to variation in smartphone model.\ignore{This indicates we cannot use these methods directly as we tend towards making such systems deploy-able in real-world scenarios.} The Kernel Ridge regression can be a strong competitor, as it is fast and easy to deploy. However the visual results suggest that although the numbers are competitive, the quality of results observed using Deep BLE based methods is much better. Also KF is infamous for being very sensitive to the calibration parameters selected. Coming up with an effective way to do that for each phone, is tedious. The DL-RNN is a strong competitor, and it performs very similar to what we see when we train LocNet using limited data. However on unseen phones, the performance of Deep BLE methods is remarkably better. Furthermore, the comparison of the 90\%ile error complemented with the Distance Error CDF comparison plot in Figure \ref{fig:box_cdf} indicate that the gap between the proposed approach and others is significant.

\section{Conclusion}
Solving localization using only RSSI signals is a challenging problem. In this paper, we formulate a new problem of coming up with a single RSSI based localization engine for different smartphones. We define different scenarios in our problem statement, considering different levels of access to the training data. We collected a large scale BLE RSSI data set for 15 different smartphones. We perform empirical experiments concluding that smartphone signals can change significantly and unpredictably across phones in different situations. We observe that in particular phones, we observe receiver failures, detecting no observations for a couple of seconds. Such a situation inevitably would lead to failure. To fix the beacon signals in such scenarios, we propose the use of \texttt{TransNet} trained using our novel statistic similarity loss (SSL) that can learn to translate beacon signals from an unknown phone RSSI to a known phone RSSI. We provide solutions for the various all the possible scenarios listed with the proposed approach. The experiments and the evaluation indicate that the proposed approach can adapt itself towards a change of smartphone models, and outperforms other state of the art methods for RSSI based localization. 

\section{Acknowledgments}

This work was supported by  Highmark, NSF (National Science Foundation) and NIDILRR (National Institute on Disability, Independent Living, and Rehabilitation Research)

\bibliographystyle{ACM-Reference-Format}
\bibliography{references}

\end{document}